\documentclass[conference]{IEEEtran}
\IEEEoverridecommandlockouts
\usepackage{cite}
\usepackage{amsmath,amssymb,amsfonts}
\usepackage{algorithmic}
\usepackage{graphicx}
\usepackage{textcomp}
\usepackage{xcolor}
\def\BibTeX{{\rm B\kern-.05em{\sc i\kern-.025em b}\kern-.08em
    T\kern-.1667em\lower.7ex\hbox{E}\kern-.125emX}}

\usepackage{microtype}
\usepackage{multirow}
\usepackage{stfloats}
\usepackage[hyphens]{url}

\newcommand{\bench}{\textsc{PilotBench}}
\newcommand{\metric}{\textsc{Pilot-Score}}
\newcommand{\numModels}{\textsc{41}}

\begin{document}

\title{PilotBench: A Benchmark for General Aviation Agents with Safety Constraints\\
}

\author{\IEEEauthorblockN{1\textsuperscript{st} Yalun Wu\textsuperscript{*}}
\IEEEauthorblockA{\textit{School of Computing} \\
\textit{National University of Singapore}\\
China \\
aaaronwww@outlook.com}
\and
\IEEEauthorblockN{2\textsuperscript{nd} Haotian Liu\textsuperscript{*}}
\IEEEauthorblockA{\textit{School of Informatics} \\
\textit{Xiamen University}\\
China \\
liuhaotian@stu.xmu.edu.cn}
\and
\IEEEauthorblockN{3\textsuperscript{rd} Zhoujun Li}
\IEEEauthorblockA{\textit{CESS} \\
\textit{Beihang University}\\
China \\
lizj@buaa.edu.cn}
\and
\IEEEauthorblockN{4\textsuperscript{th} Boyang Wang\textsuperscript{$\dagger$}}
\IEEEauthorblockA{\textit{CESS} \\
\textit{Beihang University}\\
China \\
wangboyang@buaa.edu.cn}
}
\maketitle

\begin{abstract}
As Large Language Models (LLMs) advance toward embodied AI agents operating in physical environments, a fundamental question emerges: can models trained on text corpora reliably reason about complex physics while adhering to safety constraints? We address this through \bench{}, a benchmark evaluating LLMs on safety-critical flight trajectory and attitude prediction. Built from 708 real-world general aviation trajectories spanning nine operationally distinct flight phases with synchronized 34-channel telemetry, \bench{} systematically probes the intersection of semantic understanding and physics-governed prediction through comparative analysis of LLMs and traditional forecasters. We introduce \metric{}, a composite metric balancing 60\% regression accuracy with 40\% instruction adherence and safety compliance. Comparative evaluation across 41 models uncovers a \textbf{Precision-Controllability Dichotomy}: traditional forecasters achieve superior MAE of 7.01 but lack semantic reasoning capabilities, while LLMs gain controllability with 86--89\% instruction-following at the cost of 11--14 MAE precision. Phase-stratified analysis further exposes a \textbf{Dynamic Complexity Gap}—LLM performance degrades sharply in high-workload phases such as Climb and Approach, suggesting brittle implicit physics models. These empirical discoveries motivate hybrid architectures combining LLMs' symbolic reasoning with specialized forecasters' numerical precision. \bench{} provides a rigorous foundation for advancing embodied AI in safety-constrained domains.
\end{abstract}

\begin{figure*}[b]
\centering
\includegraphics[width=\textwidth]{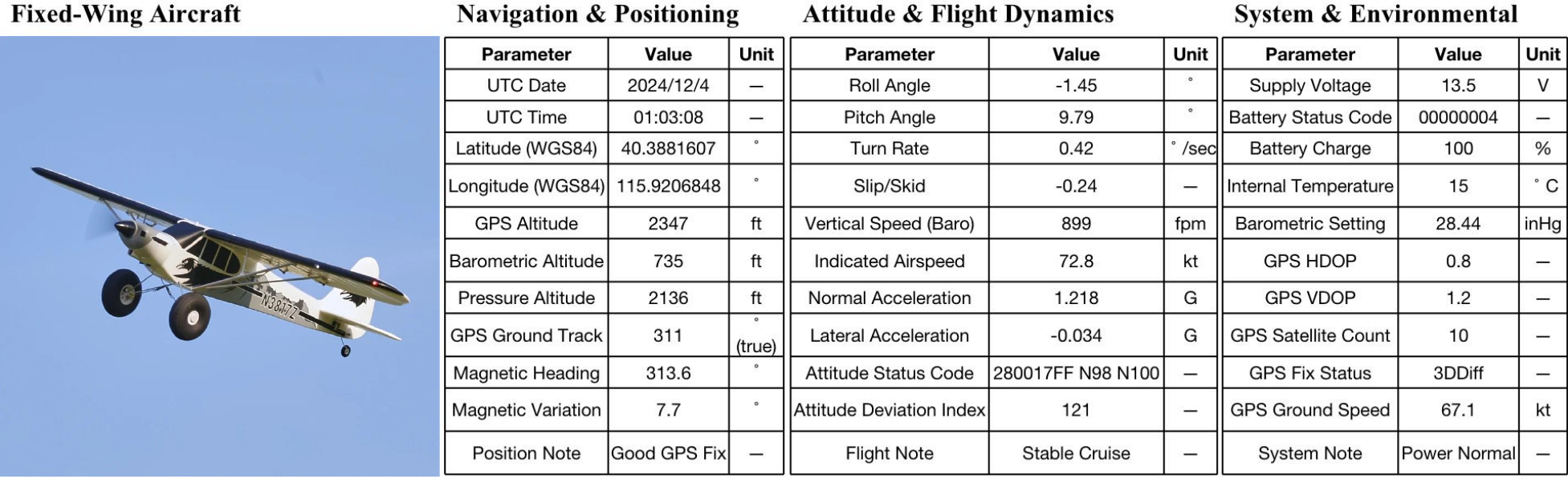}
\caption{Synchronized flight-state snapshot from \bench{} during cruise.}
\label{fig:snapshot}
\par\vspace{3pt}
\noindent\rule{\textwidth}{0.4pt}\vspace{2pt}
{\footnotesize *Equal contribution.\hspace{1.5em}$\dagger$Corresponding author.\hspace{1.5em}$^1$\url{https://github.com/haotian-io/PilotBench-A-Benchmark-for-General-Aviation-Agents-with-Safety-Constraints}}
\end{figure*}


\section{Introduction}
The rise of Large Language Models (LLMs) has sparked ambitious visions of autonomous AI agents operating in physical environments—from robotic manipulation~\cite{mon2025embodied} to autonomous driving~\cite{ma2024lampilot}. Yet a fundamental question remains unanswered: \textit{Can language models, trained primarily on text corpora, reliably reason about the complex physics governing real-world systems while adhering to strict safety constraints?}

Aviation presents an ideal testbed for this question. Flight trajectory prediction demands not only numerical accuracy but also semantic understanding of instructions, phase-aware reasoning across diverse flight regimes such as takeoff, cruise, and approach, and unwavering compliance with certified safety envelopes. Recent work demonstrates that incorporating Air Traffic Control (ATC) instructions can improve trajectory forecasting~\cite{abdulhak2024chatatc,guo2024integrating}, suggesting LLMs' potential for aviation tasks. However, existing benchmarks either focus solely on regression accuracy~\cite{liu2024research} or evaluate general reasoning without physical constraints~\cite{Zhong2023AGIEval}, leaving a critical gap: \textit{how do LLMs perform when language understanding meets physics-governed prediction in safety-critical settings?}

\begin{figure*}[b]
\centering
\includegraphics[width=.88\textwidth]{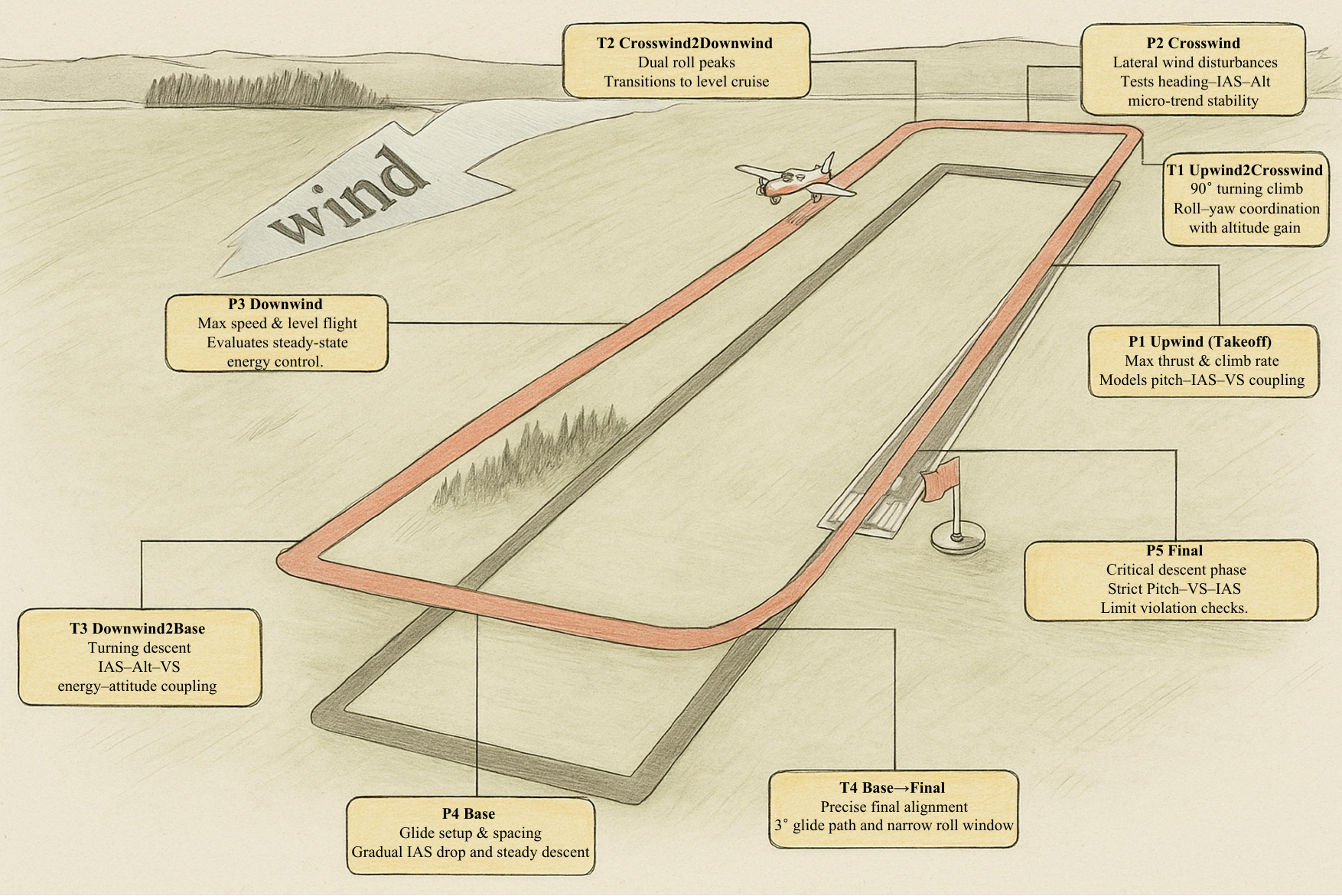}
\caption{Nine-phase segmentation in \bench{} based on standard traffic pattern.}
\label{fig:overview}
\end{figure*}

We present \bench{}, a benchmark explicitly designed to probe this intersection through empirical comparative analysis of 41 LLMs and traditional forecasting baselines on 708 real general-aviation trajectories spanning nine flight phases. We uncover a fundamental \textbf{Precision-Controllability Dichotomy}: traditional models such as FlightPatchNet achieve superior MAE of 7.01 but lack semantic reasoning, while LLMs gain 86--89\% instruction-following at the cost of 11--14 MAE precision. Our phase-stratified analysis further reveals a \textbf{Dynamic Complexity Gap}—LLMs maintain acceptable performance in steady-state phases but degrade sharply in high-dynamic regimes like climb and approach, exposing brittleness in their implicit physics models.

We formalize these findings through \metric{}, a composite metric balancing 60\% regression accuracy with 40\% instruction adherence and safety compliance, enabling holistic evaluation of aviation agents.

Our contributions: (1) \bench{}, a benchmark with 708 phase-annotated trajectories and 34-channel telemetry; (2) \metric{}, a safety-aware composite metric; and (3) empirical evidence of the Precision-Controllability Dichotomy and Dynamic Complexity Gap, motivating hybrid architectures for aviation AI.

Dataset and code are available at\textsuperscript{1}.

\section{Related Work}

Embodied AI~\cite{mon2025embodied,an2024make} requires agents to ground reasoning in physical interaction. While LLMs show promise for robotic control~\cite{wu2024cognitive} and navigation, their capacity for physics-governed prediction under safety constraints remains underexplored. In flight prediction, architectures have evolved from RNNs/LSTMs to transformer-based models such as FlightBERT++~\cite{Guo2024FlightBERTpp} and CNN--Bi-LSTM hybrids~\cite{xu2024novel,dong2024research}, with multimodal ATC integration reducing errors by over 20\%~\cite{guo2024integrating}. However, these methods treat prediction as homogeneous regression, overlooking phase differentiation and certified flight envelopes~\cite{Schimpf2024Generalised}. On the benchmark side, datasets like OpenSky~\cite{sun2025opensky} and TartanAviation~\cite{Patrikar2025TartanAviation} provide rich telemetry, while autonomous driving benchmarks such as LaMPilot~\cite{ma2024lampilot} evaluate safety-aware agents—yet aviation lacks a dedicated LLM benchmark with safety constraints. General benchmarks like AGIEval~\cite{Zhong2023AGIEval} and ToolBench~\cite{ToolLLM} also lack physical constraints. We address these gaps with \bench{}, evaluating LLMs across nine FAA-aligned flight phases, and \metric{}, quantifying adherence to certified flight envelopes.

\begin{figure*}[b]
\centering
\includegraphics[width=.75\textwidth]{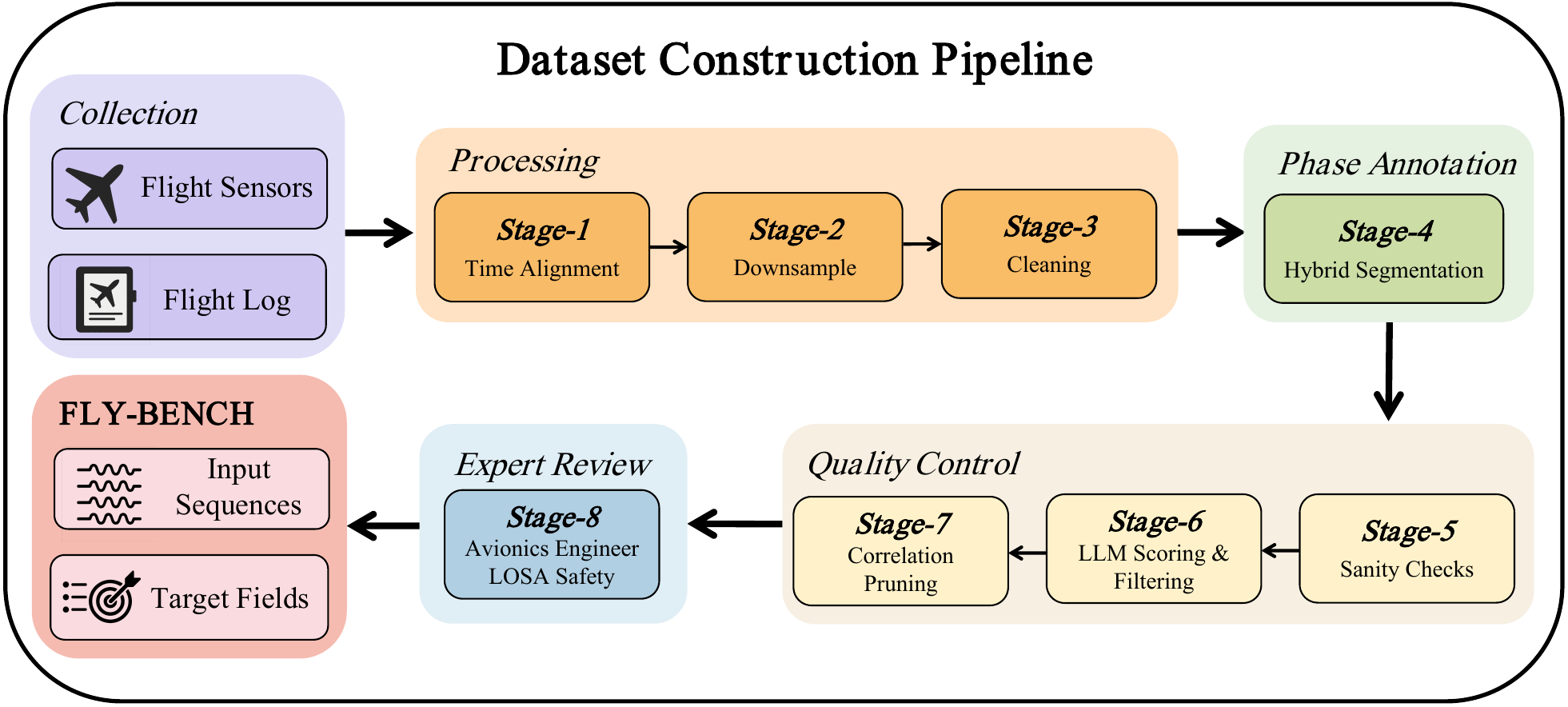}
\caption{Eight-stage pipeline for building \bench{}.}
\label{fig:pipeline}
\end{figure*}

\section{\bench{}}

The benchmark is designed to evaluate LLMs within aviation safety-critical scenarios, emphasizing structured flight segmentation, rigorous quality control, and detailed statistical evaluation of flight dynamics understanding.

\subsection{Task Definition}
\label{sec:phase_traits}

\bench{} segments flights into nine precise phases, derived from a standard aviation traffic pattern as shown in Figure~\ref{fig:overview}, consisting of five straight segments P1--P5 and four transition segments T1--T4. Phase boundaries are defined by heading change rates, roll angles, and vertical speeds, validated by annotated data.

Upwind P1 sets thrust, climb rate, VS, and IAS. T1 evaluates roll--yaw coordination through a 90° climbing turn. Crosswind P2 probes heading--IAS stability under lateral wind; T2 stresses dual-peak roll and altitude stabilisation. Downwind P3 maintains stable, level, high-speed flight parallel to the runway; T3 tests simultaneous throttle reduction, flap deployment, and descent initiation. Base P4 handles glide setup, descent rate, and IAS control; T4 demands runway alignment and glide-path interception. Final P5 enforces strict pitch, roll, and IAS limits during the critical landing approach.

\subsection{Building \bench{}}
\label{sec:build}
Inspired by existing benchmarks \cite{StableToolBench,Zhong2023AGIEval} and tailored to safety-critical aviation, \bench{} is built via an audited eight-stage pipeline, as shown in Figure~\ref{fig:pipeline}, ensuring compliance with standards.

\noindent\textbf{Data Collection.} We recorded \textbf{38\,h\,18\,min} of synchronised flight-sensor and avionics data from 31 visual-flight-rules (VFR) circuits on a DA40 and 22 dual-instruction sorties on a C172N. Each aircraft carried dual-frequency RTK-GNSS, an air-data computer (ADC), and an inertial reference unit (IRU) at 20 Hz, plus ARINC 429 streams and vendor annotations.

\noindent\textbf{Data Processing.} Channels were aligned to GNSS 1-PPS, resampled to 10 Hz, and consolidated into 34 standardised variables. ADC gaps greater than 0.8 s were forward-filled. Frames with HDOP greater than 2 or inertial residuals greater than $2.5\sigma$ were discarded, removing \textbf{1.7\%} of the corpus.

\noindent\textbf{Flight-Phase Annotation.} With 45\% missing labels, we apply rule-based seeding, contrastive boundary refinement, and expert vetting, achieving Cohen’s kappa of 0.93.

\noindent\textbf{Quality Assurance.} We apply \textbf{34} sanity checks plus automatic screening~\cite{gpt4o}; low-quality or POH-violating segments are removed, keeping MI leakage less than $0.7\%$.

\noindent\textbf{Expert Safety Review.} An avionics engineer and former LOSA auditor reviewed all segments, anonymised identifiers, and verified extreme-attitude accuracy.


\subsection{Statistics}
\label{subsec:flybench_statistics}
\begin{figure*}[ht]
\centering
\includegraphics[width=\textwidth]{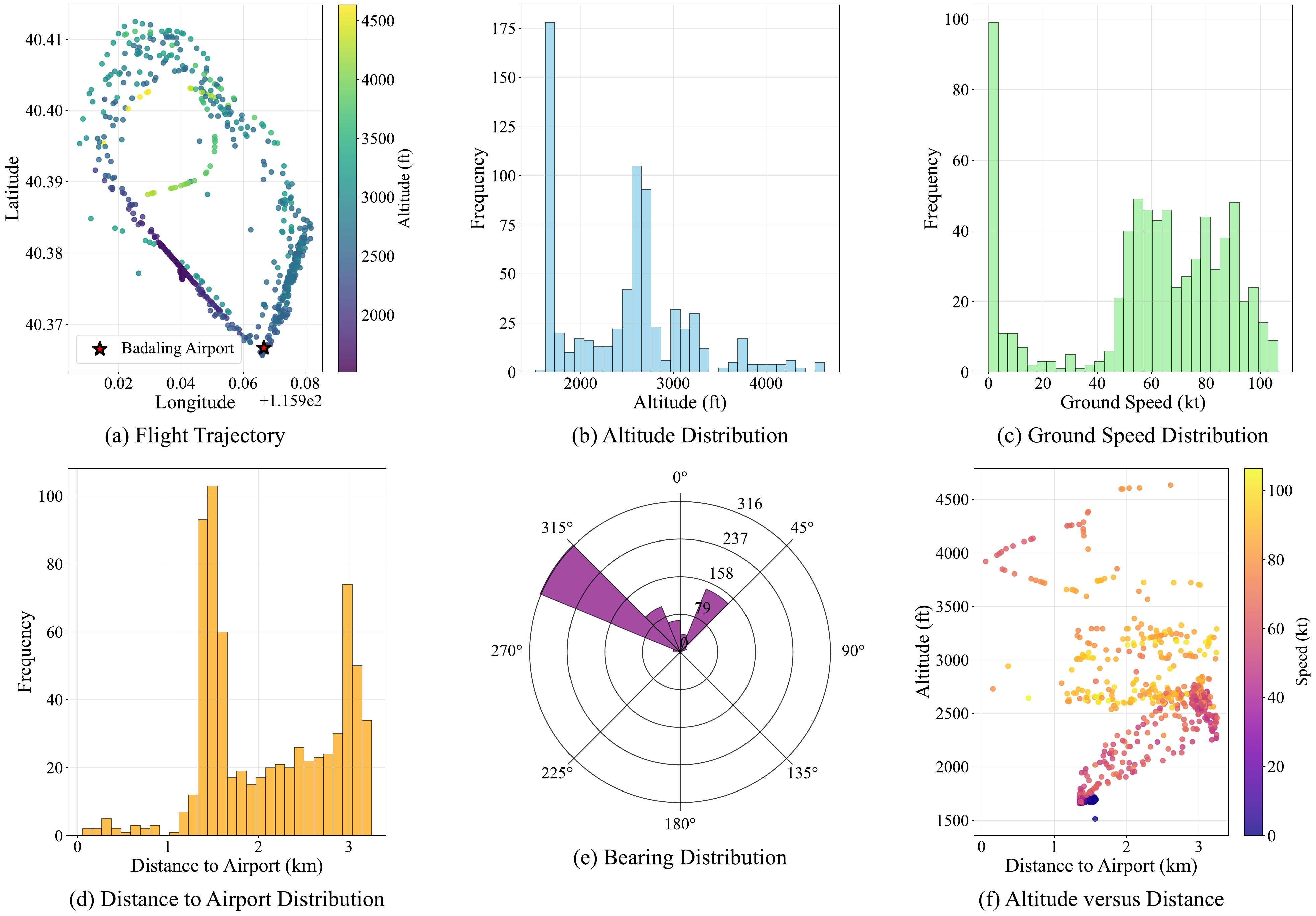}
\caption{Flight and spatial statistics of \bench{}.}
\label{fig:flybench_overview}
\vspace{-5mm}
\end{figure*}

The \bench{} corpus contains 708 trajectory segments from a ten-day campaign, spanning altitudes 1,514--4,633\,ft with a mean of 2,485\,ft MSL, ground speeds 0--106.4\,kt averaging 58.5\,kt, with 34 synchronised sensor features per record. Figure~\ref{fig:flybench_overview} summarises key distributions: sorties trace a 4\,km oval pattern with two dominant altitude bands at 1.7--2.1\,kft and 2.4--3.2\,kft covering 70\%, three speed clusters, bimodal distance-to-airport peaks at 1.3 and 2.8\,km, and headings concentrated at 270°--315° consistent with counter-clockwise circuits.

\subsection{\metric{}}
The metric $\mathcal{F}: \mathbb{R}^+ \times [0,1]^3 \to [0,100]$ combines regression accuracy and instruction-following:
\begin{equation}
\mathcal{F}(\text{MAE}, \text{RMSE}, \mathbf{c}) = 0.6 \cdot \mathcal{R}(\text{MAE}, \text{RMSE}) + 0.4 \cdot \mathcal{I}(\mathbf{c})
\end{equation}
where $\mathcal{R} = 0.6 \cdot S_{\text{MAE}} + 0.4 \cdot S_{\text{RMSE}}$ via piecewise linear scoring and $\mathcal{I}(\mathbf{c}) = 0.5 c_1 + 0.3 c_2 + 0.2 c_3$ over Field Completeness, Validity, and Format. Instruction-following is computed by automated parsers; a floor score of 5 prevents complete failure classification. The metric is monotonically decreasing in error and increasing in compliance, bounded in $[5,100]$, and aligned with Required Navigation Performance (RNP) standards: $\mathcal{F}\!\geq\!90$ maps to RNP~0.1, $\geq\!70$ to RNP~0.3--1.0, $\geq\!50$ to RNP~2--4~\cite{federal2022airplane}. The 60/40 regression-instruction weighting follows aviation safety assessments~\cite{spence2015international,boyd2017review,puranik2017energy}.

\section{Experiment}

We evaluate \numModels{} LLMs on \bench{}. Each model ingests structured historical waypoints and outputs one-step trajectory and attitude predictions in a fixed schema. Decoding is deterministic with temperature 0 and $p$=1.0. Experiments ran on eight NVIDIA A100 80GB GPUs; the model roster spans Qwen, DeepSeek, GLM, InternLM, GPT, and Doubao.

\begin{table*}[!htbp]
\caption{Main results on \bench{}}
\begin{center}
\scriptsize
\resizebox{\textwidth}{!}{%
\begin{tabular}{|l|c|c|c|c|c|c|c|c|}
\hline
& \multicolumn{3}{c|}{\textbf{Regression Error}} & \multicolumn{4}{c|}{\textbf{Instruction Following}} & \multirow{2}{*}{\textbf{\metric{}}} \\
\cline{2-8}
\textbf{Model} & \textbf{MAE $\downarrow$} & \textbf{RMSE $\downarrow$} & \textbf{Overall $\uparrow$} & \textbf{Field Comp.} & \textbf{Field Valid} & \textbf{Format} & \textbf{Overall} & \\
\hline
Qwen3-32B & \textbf{9.5409} & 14.7005 & 85.43 & 100.0 & 91.5 & 100.0 & \textbf{97.43} & \textbf{90.23} \\
Qwen2.5-72B-Instruct & 11.9144 & 16.3591 & 83.20 & 100.0 & 88.8 & 100.0 & 96.64 & 88.58 \\
DeepSeek-V3 & 11.9410 & 16.3006 & 83.19 & 100.0 & 86.7 & 100.0 & 96.00 & 88.31 \\
GPT-4o-mini & 12.1116 & 16.5832 & 82.99 & 99.8 & 86.3 & 99.8 & 95.74 & 88.09 \\
GPT-o3-mini & 10.9403 & 15.0377 & 84.24 & 100.0 & 76.8 & 100.0 & 93.05 & 87.76 \\
Doubao-1.5Pro-32k & 11.6238 & 15.8508 & 83.53 & 100.0 & 78.3 & 100.0 & 93.50 & 87.52 \\
LoRA-Qwen2.5-14B-Instruct & 13.1038 & 31.9880 & 79.12 & 99.8 & 89.4 & 99.8 & 96.67 & 86.14 \\
Qwen2.5-32B-Instruct & 13.0683 & 32.7015 & 79.01 & 100.0 & 82.2 & 100.0 & 94.66 & 85.27 \\
QwQ-32B & 9.7241 & 13.4456 & 85.53 & 100.0 & 46.5 & 100.0 & 83.95 & 84.90 \\
Qwen2.5-14B-Instruct & 14.0535 & 42.2590 & 76.31 & 99.8 & 89.3 & 99.8 & 96.65 & 84.44 \\
Qwen2.5-VL-72B-Instruct & 20.3666 & 76.1834 & 65.66 & 100.0 & 83.0 & 100.0 & 94.89 & 77.35 \\
Qwen2.5-7B-Instruct & 35.5291 & 110.1615 & 54.98 & 100.0 & 84.4 & 100.0 & 95.32 & 71.11 \\
GLM-4-9B & 41.1249 & 134.9971 & 50.75 & 100.0 & 88.2 & 100.0 & 96.45 & 69.03 \\
DeepSeek-R1-Distill-Llama-70B & 37.1562 & 129.6232 & 52.77 & 99.9 & 44.6 & 100.0 & 83.34 & 65.00 \\
QVQ-72B-Preview & 9.7267 & \textbf{11.2431} & \textbf{85.97} & 2.7 & \textbf{97.7} & 2.7 & 30.67 & 63.85 \\
DeepSeek-R1-Distill-Qwen-32B & 68.8524 & 190.3993 & 38.24 & 100.0 & 70.3 & 100.0 & 91.07 & 59.37 \\
Qwen3-14B & 150.4770 & 293.6560 & 19.84 & 100.0 & 80.3 & 100.0 & 94.09 & 49.54 \\
DeepSeek-R1-Distill-Qwen-7B & 273.8185 & 422.7081 & 6.57 & 99.5 & 82.6 & 99.7 & 94.42 & 41.71 \\
GLM-Z1-Rumination-32B & 84.5391 & 224.6919 & 32.23 & 29.8 & 25.2 & 99.0 & 42.03 & 36.15 \\
GLM-4-9B-Chat & 483.7880 & 553.1207 & 5.00 & 100.0 & 0.7 & 100.0 & 70.21 & 31.08 \\

\hline
\end{tabular}}
\vspace{-6mm}
\label{tab:flybench_results}
\end{center}
\end{table*}

Results (Table~\ref{tab:flybench_results}) show \textbf{Qwen3-32B} achieves the best MAE of 9.54. Scaling helps but non-linearly; architecture and training matter more than size. \textsc{QVQ-72B-Preview} is numerically accurate yet often refuses instructions, limiting utility. Large models like Qwen2.5-72B reach 88.8\% field validity; small models such as GLM-4-9B-Chat maintain format but attain only 0.7\% validity, decoupling syntax from semantics. Errors are lowest in cruise, moderate in descent, and highest in climb, motivating phase-aware modeling.

Key failure patterns include numerical anomalies, flight-logic errors, structural issues, refusal responses, and temporal discontinuities—informing hybrid designs where LLMs handle intent while specialized modules ensure physics consistency.

\section{Performance Evaluation Across Methods and Flight Phases}

To complement the 41-model screening, we conduct an ablation on 9 configurations: 3 traditional baselines, namely FlightPatchNet, DLinear, and PatchTST, and 6 LLM variants covering GPT-4o and Qwen3-32B under varied prompting strategies.

\subsection{The Precision-Controllability Dichotomy}

\begin{figure}[!htbp]
\centering
\includegraphics[width=0.95\columnwidth]{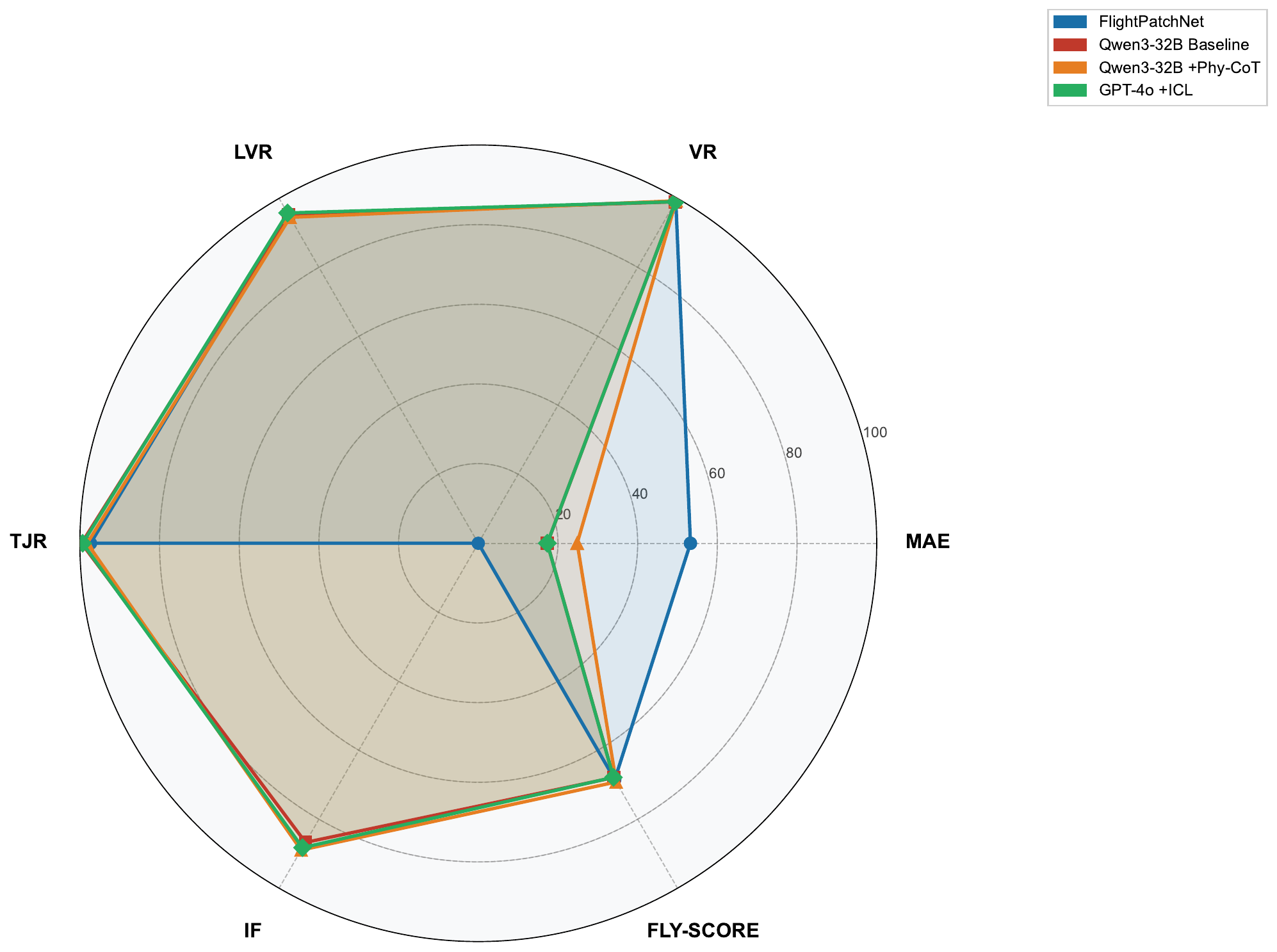}
\caption{Performance radar: traditional models shown in blue dominate MAE/VR; LLMs shown in orange, green, and purple gain IF controllability. Qwen3-32B +Phy-CoT achieves the best balance.}
\label{fig:radar}
\end{figure}

Figure~\ref{fig:radar} illustrates a fundamental trade-off in current modeling paradigms. Traditional forecasting baselines excel in pure regression with FlightPatchNet achieving MAE=7.01, leveraging inductive biases tailored for time-series extrapolation. However, their inability to process natural language renders them inert in human-on-the-loop scenarios requiring semantic guidance. 

Conversely, LLMs sacrifice profound numerical precision with MAE ranging 11.28--13.59 to gain semantic controllability. This \textbf{Precision-Controllability Dichotomy} forces a choice: traditional models for fixed-trajectory prediction versus LLMs for adaptive, instruction-guided maneuvering. Notably, Qwen3-32B with Phy-CoT achieves the most effective compromise at \metric{}=69.21, maintaining acceptable regression error while enabling high-level safety constraints with VR below 0.8\% and instruction adherence IF=88.9.

\subsection{Safety and Accuracy as Orthogonal Targets}

\begin{figure}[!htbp]
\centering
\includegraphics[width=0.98\columnwidth]{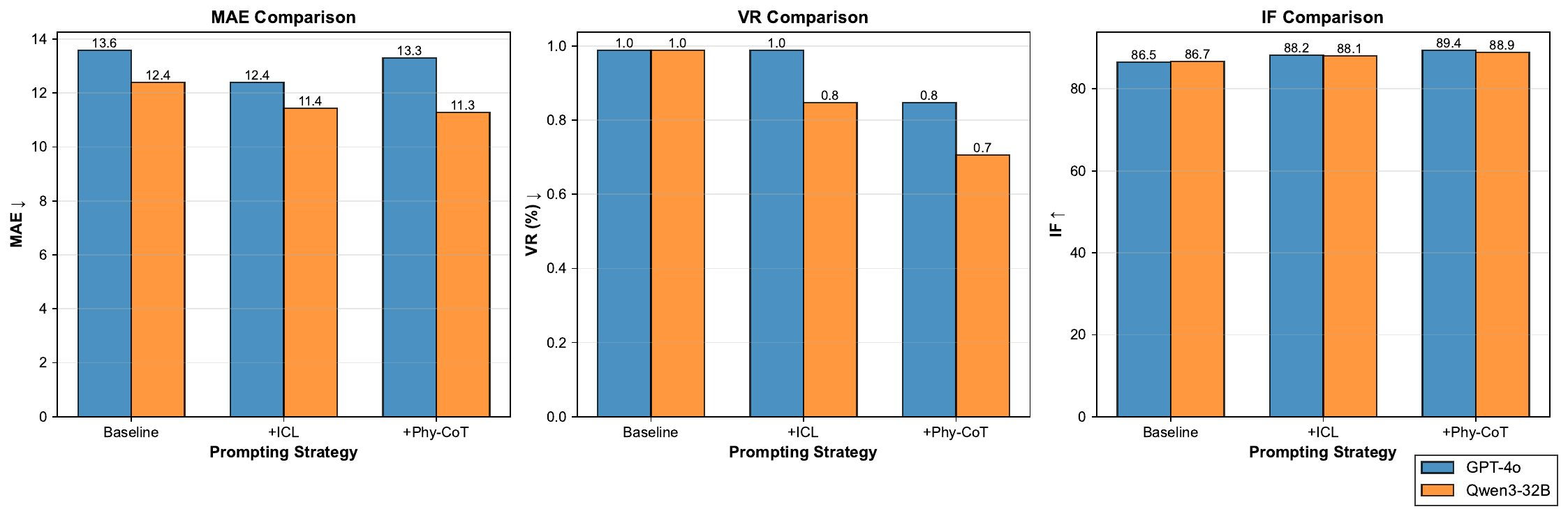}
\caption{Decoupling prompting effects: +ICL benefits MAE precision; +Phy-CoT drives VR safety gains, suggesting distinct mechanisms for pattern matching vs.\ constraint satisfaction.}
\label{fig:prompting}
\end{figure}

Figure~\ref{fig:prompting} dissects the mechanisms of prompting improvements, revealing that accuracy and safety are improved via distinct pathways. 
\begin{itemize}
    \item \textbf{In-Context Learning (+ICL)} primarily acts as a \textit{syntax stabilizer}. By providing exemplars, it reduces MAE—for instance, achieving -1.19 improvement for GPT-4o—and improves format compliance, but offers limited gains in physical safety.
    \item \textbf{Physics-CoT (+Phy-CoT)} functions as a \textit{semantic constraint injector}. It explicitly grounds the reasoning process in flight dynamics, yielding dramatic safety improvements with VR reduction of approximately 25\% that ICL alone cannot achieve.
\end{itemize}
This suggests that while few-shot prompting suffices for pattern matching, reliable safety-critical behavior requires explicit reasoning chains that verify physical feasibility before output generation.

\subsection{The Dynamic Complexity Gap}

\begin{figure}[!htbp]
\centering
\includegraphics[width=0.95\columnwidth]{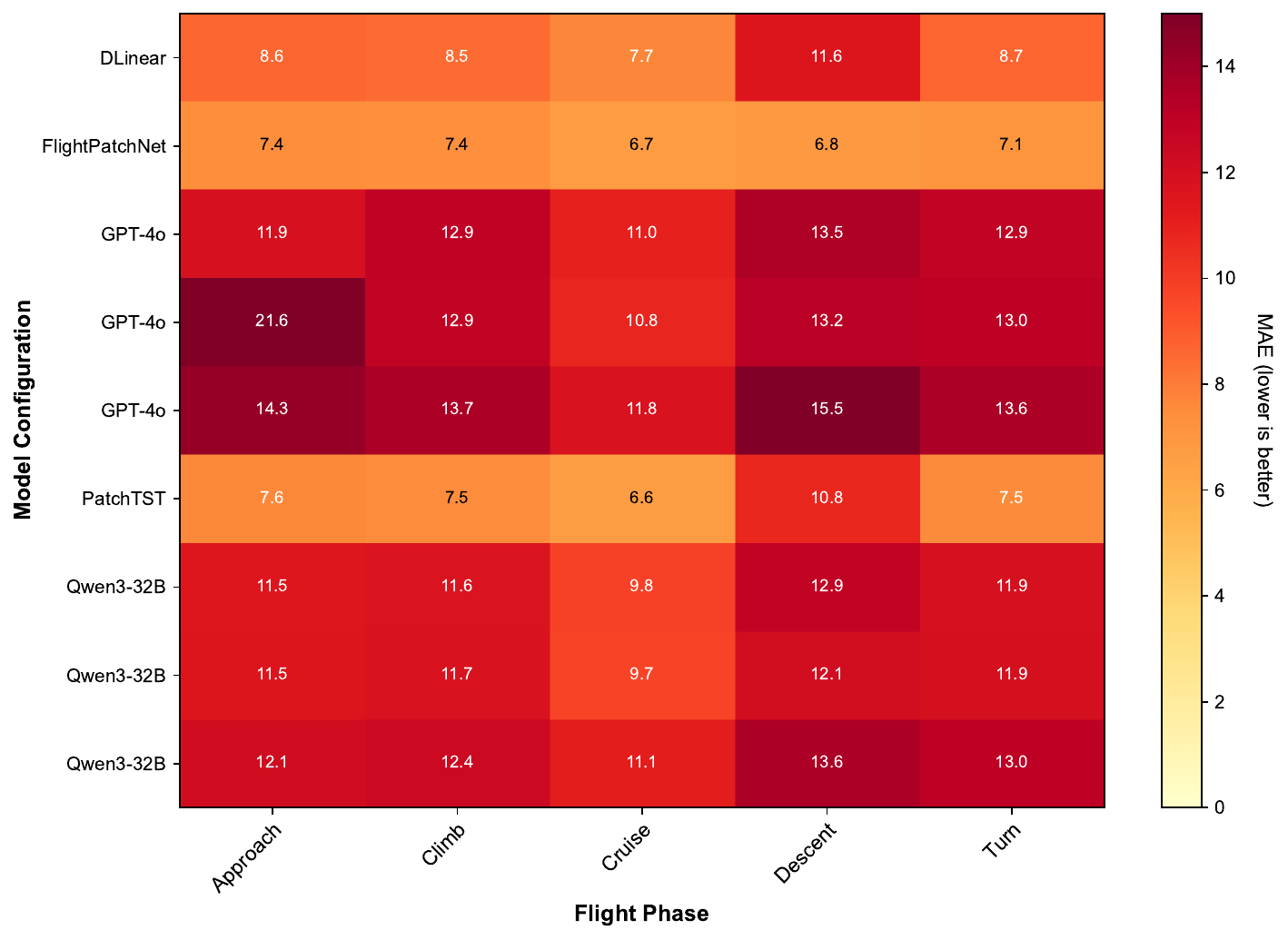}
\caption{MAE heatmap across flight phases. LLM errors spike in Climb and Approach, marked in red, while traditional baselines remain stable in yellow, exposing the Dynamic Complexity Gap.}
\label{fig:heatmap}
\end{figure}

Phase-stratified analysis (Figure~\ref{fig:heatmap}) exposes a \textbf{Dynamic Complexity Gap}. While all models perform robustly in low-dynamic phases like Cruise and Descent with MAE of 7--10, LLM performance degrades disproportionately in high-workload phases.

In Climb and Approach---characterized by coupled changes in altitude, airspeed, and configuration---LLM error rates spike above MAE 13, whereas specialized forecasters like FlightPatchNet maintain stability below MAE 8. This sensitivity suggests that LLMs' implicit physics models are brittle: adequate for steady-state extrapolation but fragile under the coupled non-linear dynamics of aggressive maneuvering.

\section{Conclusion}

\bench{} reveals that LLMs demonstrate emergent capabilities in semantic interpretation and instruction following, yet face intrinsic limitations in high-precision physics modeling. The \textbf{Precision-Controllability Dichotomy} exposes that traditional forecasters excel at numerical prediction but remain inert to natural language guidance, while LLMs sacrifice regression fidelity to gain semantic controllability. The \textbf{Dynamic Complexity Gap} reveals that LLMs' acceptable performance in steady-state regimes collapses under coupled, high-workload maneuvers.

These findings motivate a \textbf{Hybrid Architectural Paradigm}: leveraging LLMs for intent understanding while delegating high-frequency control to specialized forecasters, generalizing beyond aviation to any safety-critical domain requiring both semantic reasoning and physical prediction.



\section*{Acknowledgment}
This work was supported in part by the National Natural Science Foundation of China (Grant Nos. 62276017, 62406033, U1636211, 61672081), and the State Key Laboratory of Complex \& Critical Software Environment (Grant No.\ SKLCCSE-2024ZX-18). This study used exclusively anonymized device-level flight sensor data; no human subjects or personally identifiable information were involved.

\noindent\textbf{AI Usage Disclosure.} In accordance with IEEE policy, we disclose that AI-assisted writing tools (e.g., GPT-5) were used for language refinement and editing in the preparation of this manuscript. The use of AI-generated text is limited to improving clarity and readability. No AI tools were used in the design of the study, data collection, data analysis, or the derivation of scientific conclusions.

\bibliographystyle{IEEEtran}
\bibliography{refs}

\end{document}